# Look at Boundary: A Boundary-Aware Face Alignment Algorithm

Wayne Wu [*,1,2], Chen Qian[2], Shuo Yang[3], Quan Wang[2], Yici Cai[1], Qiang Zhou[1]

[1]Tsinghua National Laboratory for Information Science and Technology (TNList),
Department of Computer Science and Technology, Tsinghua University
[2]SenseTime Research
[3]Amazon Rekognition

[1]`wwy15@mails.tsinghua.edu.cn` [1]`caiyc@mail.tsinghua.edu.cn` [1]`zhouqiang@tsinghua.edu.cn`
[2]`{qianchen, wangquan}@sensetime.com` [3]`shuoy@amazon.com`

## Abstract

*We present a novel boundary-aware face alignment algorithm by utilising boundary lines as the geometric structure of a human face to help facial landmark localisation. Unlike the conventional heatmap based method and regression based method, our approach derives face landmarks from boundary lines which remove the ambiguities in the landmark definition. Three questions are explored and answered by this work: 1. Why using boundary? 2. How to use boundary? 3. What is the relationship between boundary estimation and landmarks localisation? Our boundary-aware face alignment algorithm achieves 3.49% mean error on 300-W Fullset, which outperforms state-of-the-art methods by a large margin. Our method can also easily integrate information from other datasets. By utilising boundary information of 300-W dataset, our method achieves 3.92% mean error with 0.39% failure rate on COFW dataset, and 1.25% mean error on AFLW-Full dataset. Moreover, we propose a new dataset WFLW to unify training and testing across different factors, including poses, expressions, illuminations, makeups, occlusions, and blurriness. Dataset and model will be publicly available at* [https://wywu.github.io/projects/LAB/LAB.html](https://wywu.github.io/projects/LAB/LAB.html)

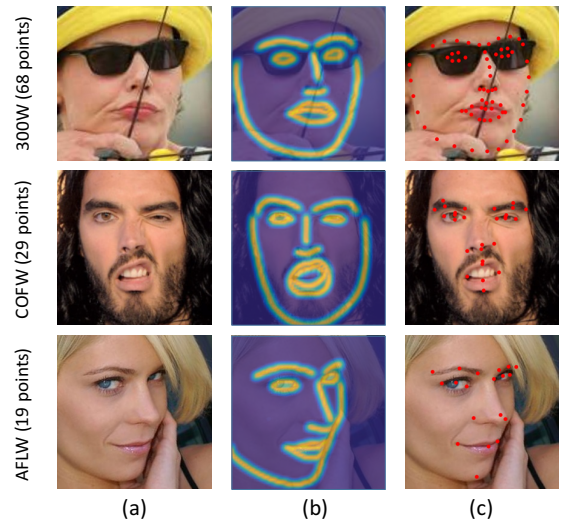

Figure 1: The first column shows the face images from different datasets with different number of landmarks. The second column illustrates the universally defined facial boundaries estimated by our methods. With the help of boundary information, our approach achieves high accuracy localisation results across multiple datasets and annotation protocols, as shown in the third column.

## 1. Introduction

Face alignment, which refers to facial landmark detection in this work, serves as a key step for many face applications, e.g., face recognition [75], face verification [48, 49] and face frontalisation [21]. The objective of this paper is to devise an effective face alignment algorithm to handle faces with unconstrained pose variation and occlusion across multiple datasets and annotation protocols.

Different to face detection [45] and recognition [75], face alignment identifies geometry structure of human face which can be viewed as modeling highly structured output. Each facial landmark is strongly associated with a well-defined facial boundary, e.g., eyelid and nose bridge. However, compared to boundaries, facial landmarks are not so well-defined. Facial landmarks other than corners can hardly remain the same semantical locations with large pose variation and occlusion. Besides, different annotation schemes of existing datasets lead to a different number of landmarks [28, 5, 66, 30] (19/29/68/194 points) and annotation scheme of future face alignment datasets can hardly be determined. We believe the reasoning of a unique facial

---

[*]This work was done during an internship at SenseTime Research.

structure is the key to localise facial landmarks since human face does not include ambiguities.

To this end, we use well-defined facial boundaries to represent the geometric structure of the human face. It is easier to identify facial boundaries comparing to facial landmarks under large pose and occlusion. In this work, we represent facial structure using 13 boundary lines. Each facial boundary line can be interpolated from a sufficient number of facial landmarks across multiple datasets, which will not suffer from inconsistency of the annotation schemes.

Our boundary-aware face alignment algorithm contains two stages. We first estimate facial boundary heatmaps and then regress landmarks with the help of boundary heatmaps. As noticed in Fig. 1, facial landmarks of different annotation schemes can be derived from boundary heatmaps with the same definition. To explore the relationship between facial boundaries and landmarks, we introduce adversarial learning ideas by using a landmark-based boundary effectiveness discriminator. Experiments have shown that the better quality estimated boundaries have, the more accurate landmarks will be. The boundary heatmap estimator, landmark regressor, and boundary effectiveness discriminator can be jointly learned in an end-to-end manner.

We used stacked hourglass structure [35] to estimate facial boundary heatmap and model the structure between facial boundaries through message passing [11, 63] to increase its robustness to occlusion. After generating facial boundary heatmaps, the next step is deriving facial landmarks using boundaries. The boundary heatmaps serve as structure cue to guide feature learning for the landmark regressor. We observe that a model guided by ground truth boundary heatmaps can achieve $76.26\%$ AUC on 300W [39] test while the state-of-the-art method [15] can only achieve $54.85\%$. This suggests the richness of information contained in boundary heatmaps. To fully utilise the structure information, we apply boundary heatmaps at multiple stages in the landmark regression network. Our experiment shows that the more stages boundary heatmaps are used in feature learning, the better landmark prediction results we will get.

We evaluate the proposed method on three popular face alignment benchmarks including 300W [39], COFW [5], and AFLW [28]. Our approach significantly outperforms previous state-of-the-art methods by a large margin. $3.49\%$ mean error on 300-W Fullset, $3.92\%$ mean error with $0.39\%$ failure rate on COFW and $1.25\%$ mean error on AFLW-Full dataset respectively. To unify the evaluation, we propose a new large dataset named Wider Facial Landmarks in-the-wild (WFLW) which contain $10,000$ images. Our new dataset introduces large pose, expression, and occlusion variance. Each image is annotated with $98$ landmarks and $6$ attributes. Comprehensive ablation study demonstrates the effectiveness of each component.

## 2. Related Work

In the literature of face alignment, besides classic methods (ASMs [34, 23], AAMs [13, 41, 33, 25], CLMs [29, 42] and Cascaded Regression Models [7, 5, 58, 8, 72, 73, 18]), recently, state-of-the-art performance has been achieved with Deep Convolutional Neural Networks (DCNNs). These methods mainly fall into two categories, *i.e.*, coordinate regression model and heatmap regression model.

**Coordinate regression models** directly learn the mapping from the input image to the landmark coordinates vector. Zhang *et al.* [70] frames the problem as a multi-task learning problem, learns landmark coordinates and predicts facial attributes at the same time. MDM [51] is the first end-to-end recurrent convolutional system for face alignment from coarse to fine. TSR [31] splits face into several parts to ease the parts variations and regresses the coordinates of different parts respectively. Even though coordinate regression models have the advantage of explicit inference of landmark coordinates without any post-processing. Nevertheless, they are not performing as well as heatmap regression models.

**Heatmap regression models**, which generate likelihood heatmaps for each landmark respectively, have recently achieved state-of-the-art performance in face alignment. CALE [4] is a two-stage convolutional aggregation model to aggregate score maps predicted by detection stage along with early CNN features for final heatmap regression. Yang *et al.* [60] uses a two parts network, *i.e.*, a supervised transformation to normalise faces and a stacked hourglass network [35] to get prediction heatmaps. Most recently, JMFA [15] achieves state-of-the-art accuracy by leveraging stacked hourglass network [35] for multi-view face alignment and demonstrates better than the best three entries of the last Menpo Challenge [66].

Since boundary detection was set as one of the most fundamental problems in computer vision and there have emerged a large number of materials [56, 52, 44, 65, 43]. It has been proved efficient in vision tasks as segmentation [32, 27, 22] and object detection [36, 50, 37]. In face alignment, boundary information demonstrates especial importance because almost all of the landmarks are defined lying on the facial boundaries. However, as far as we know, in face alignment task, no work before has investigated the use of boundary information from an *explicit* perspective.

The recent advance in human pose estimation partially inspires our method of boundary heatmaps estimation. Stacked hourglass network [35] achieves compelling accuracy with a bottom-up, top-down design which endows the network with capabilities of obtaining multi-scale information. Message passing [11, 63] has shown great power in structure modeling of human joints. Recently, adversarial learning [9, 10] is adopted to further improve the accuracy of estimated human pose under heavy occlusion.

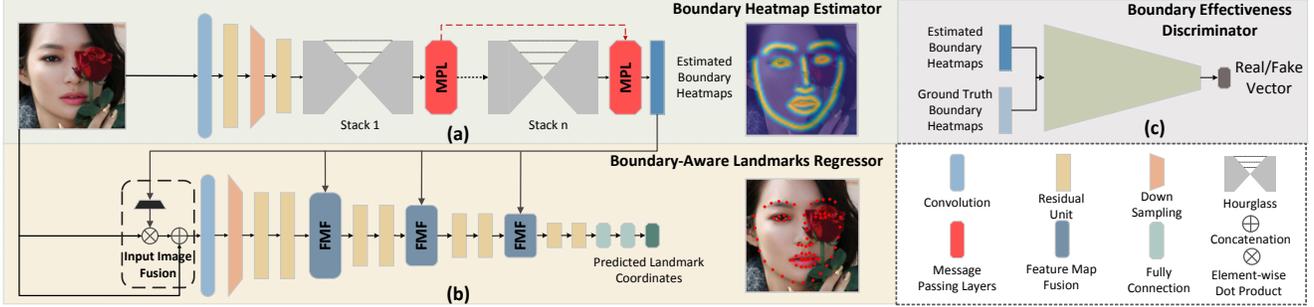

Figure 2: Overview of our Boundary-Aware Face Alignment framework. (a) Boundary heatmap estimator, which based on hourglass network is used to estimate boundary heatmaps. Message passing layers are introduced to handle occlusion. (b) Boundary-aware landmarks regressor is used to generate the final prediction of landmarks. Boundary heatmap fusion scheme is introduced to incorporate boundary information into the feature learning of regressor. (c) Boundary effectiveness discriminator, which distinguishes "real" boundary heatmaps from "fake", is used to further improve the quality of the estimated boundary heatmaps.

## 3. Boundary-Aware Face Alignment

As mentioned in the introduction, landmarks have difficulty in presenting accurate and universal geometric structure of face images. We propose facial boundary as geometric structure representation and help landmarks regression problem in the end. Boundaries are detailed and well-defined structure descriptions, which are consistent across head poses and datasets. They are also closely related to landmarks since most of the landmarks are located along boundary lines.

Other choices are also available for geometric structure representations. Recent works [31, 47, 19] has adopted facial parts to aid face alignment tasks. However, facial parts are too coarse thus not as powerful as boundary lines. Another choice would be face parsing results. Face parsing leads to disjoint facial components which needs the boundaries of each component form a closed loop. However, some facial organs such as nose are naturally blended into the whole face thus are inaccurate to be defined as separate parts. On the contrary, boundary lines are not necessary to form a closed loop, which is more flexible in representing geometric structure. Experiments in Sec 4.2 have shown that boundary lines are the best choice to aid landmark coordinates regression.

The detailed configuration of our proposed Boundary-Aware Face Alignment framework is illustrated in Fig. 2. It is composed of three closely related components: Boundary-Aware Landmark Regressor, Boundary Heatmap Estimator and Landmark-Based Boundary Effectiveness Discriminator. Boundary-Aware Landmark Regressor incorporates boundary information in a multi-stage manner to predict landmark coordinates. Boundary Heatmap Estimator produces boundary heatmaps as face geometric structure. Since boundary information is used heavily, the quality of boundary heatmaps is crucial for final landmark regression. We introduce adversarial learning idea [20] by proposing Landmark-Based Boundary Effectiveness Dis-criminator, which is paired with the Boundary Heatmap Estimator. This discriminator can further improve the quality of boundary heatmaps and lead to better landmark coordinates prediction.

### 3.1. Boundary-aware landmarks regressor

In order to fuse boundary line into feature learning, we transform landmarks to boundary heatmaps to aid the learning of feature. The responses of each pixel in boundary heatmap are decided by its distance to the corresponding boundary line. As shown in Fig. 3, the details of boundary heatmap are defined as follows.

Given a face image $I$, denote its ground truth annotation by $L$ landmarks as $S = \{s_l\}_{l=1}^{L}$. $K$ subsets $S_i \subset S$ are defined to represent landmarks belongs to $K$ boundaries respectively, such as upper left eyelid and nose bridge. For each boundary, $S_i$ is interpolated to get a dense boundary line. Then a binary boundary map $B_i$, the same size as $I$, is formed by setting only points on the boundary line to be 1, others 0. Finally, a distance transform is performed based on each $B_i$ to get distance map $D_i$. We use a gaussian expression with standard deviation $\sigma$ to transform the distance map to ground-truth boundary heatmap $M_i$. $3\sigma$ is used to threshold $D_i$ to make boundary heatmaps focus more on boundary areas. In practice, the length of the ground-truth boundary heatmap side is set to a quarter of the size of $I$ for computation efficiency.

$$M_i(x,y) = \begin{cases} \exp(-\frac{D_i(x,y)^2}{2\sigma^2}), & \text{if } D_i(x,y) < 3\sigma \\ 0, & \text{otherwise} \end{cases} \quad (1)$$

In order to fully utilise the rich information contained in boundary heatmaps, we propose a multi-stage boundary heatmap fusion scheme. As illustrated in Fig. 2, A four-stage res-18 network is adopted as our baseline network. Boundary heatmap fusion is conducted at the input and every stage of the network. Comprehensive results in Sec. 4.2

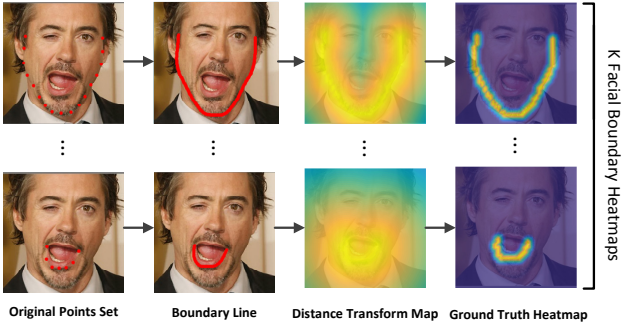

Original Points Set    Boundary Line    Distance Transform Map    Ground Truth Heatmap

Figure 3: An illustration of the process of ground truth heatmap generation. Each row represents the process of one specific facial boundary, *i.e.*, facial outer contour, left eyebrow, right eyebrow, nose bridge, nose boundary, left/right upper/lower eyelid and upper/lower side of upper/lower lip.

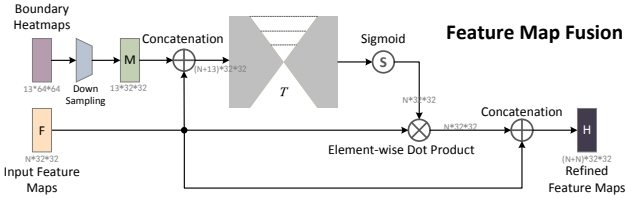

Figure 4: An illustration of the *feature map fusion* scheme. Boundary cues and input feature maps are fused together to get a refined feature with the usage of a hourglass module.

have shown that the more fusion we conducted to the baseline network, the better performance we can get.

**Input image fusion.** To fuse boundary heatmap $M$ with input image $I$, the fused input $H$ is defined as:

$$H = I \oplus (M_1 \otimes I) \oplus ... \oplus (M_T \otimes I) \quad (2)$$

where $\otimes$ represents the element-wise dot product operation and $\oplus$ represents channel-wise concatenation. The above design makes fused input focus only on detailed texture around boundaries. Thus most background and texture-less face regions are ignored which greatly enhance the effectiveness of input. The original input is also concatenated to the fused ones to keep other valuable information in the original image.

**Feature map fusion.** Similar to above, to fuse boundary heatmap $M$ with feature map $F$, the fused feature map $H$ is defined as:

$$H = F \oplus (F \otimes T(M \oplus F)) \quad (3)$$

Since the number of channels of $M$ equals to the number of pre-defined boundaries, which is constant. A transform function $T$ is necessary to convert $M$ to have the same channels with $F$. We choose hourglass structure subnet as $T$ to keep feature map size. Down-sampling and up-sampling are performed symmetrically. Skip connections are used to combine multi-scale information. Then a sigmoid layer normalises the output range to $[0, 1]$. Another simple choice would be consecutive convolutional layers with stride equals to one, which covers relatively local areas. Experiments in Sec. 4.2 have demonstrated the superiority of hourglass structure. Details of feature map fusion subnet are illustrated in Fig. 4.

Since boundary heatmaps are used heavily in landmarks coordinates regression. The quality of boundary heatmaps is essential to the prediction accuracy. By fusing ground truth boundary heatmaps, our method can achieve $76.26\%$ AUC on 300-W test, comparing to the state-of-art result $54.85\%$. Based on this experiment, in the following sections, several methods will be introduced to improve the quality of generated boundary heatmaps. Experiment in ablation study also shows the consistent performance gain with better heatmap quality.

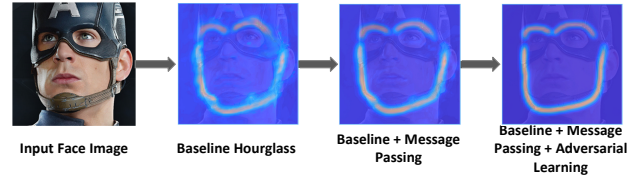

Input Face Image    Baseline Hourglass    Baseline + Message Passing    Baseline + Message Passing + Adversarial Learning

Figure 5: An illustration of the effectiveness of message passing and adversarial learning. With the message passing and adversarial learning addition, the quality of the estimated boundary is well improved to be more and more plausible and focused.

### 3.2. Boundary heatmap estimator

Following previous work in face alignment [15, 60] and human pose [35, 12, 62], we use stacked hourglass as the baseline of boundary heatmap estimator. Mean square error (MSE) between generated and groundtruth boundary heatmaps is optimized. However, as demonstrated in Fig. 5, when heavy occlusions happen, the generated heatmaps always suffer from the noisy and multi-mode response, which has also been mentioned in [9, 12].

In order to relieve the problem caused by occlusion, we introduce message passing layers to pass information between boundaries. This process is visualised in Fig. 6. During occlusion, visible boundaries can provide help to occluded ones according to face structure. **Intra-level message passing** is used at the end of each stack to pass information between different boundary heatmaps. Thus, information can be passed from visible boundaries to occluded ones. Moreover, since different stacks of hourglass focus on different aspects of face information. **Inter-level message passing** is adopted to pass message from lower stacks to the higher stacks to keep the quality of boundary heatmaps when stacking more hourglass subnets.

We implemented message passing following [11]. In this implementation, the feature map at the end of each stack needs be divided into $K$ branches, where $K$ is the number of boundaries, each represents a type of boundary feature map. This requirement demonstrates the advantage of our boundary heatmaps compared with landmark heatmaps [15, 60] for the small and constant number $K$ of them. Thus, the computational and parameter cost of message passing layers within boundaries is small while it is not practical for message passing within 68 or even 194 landmarks.

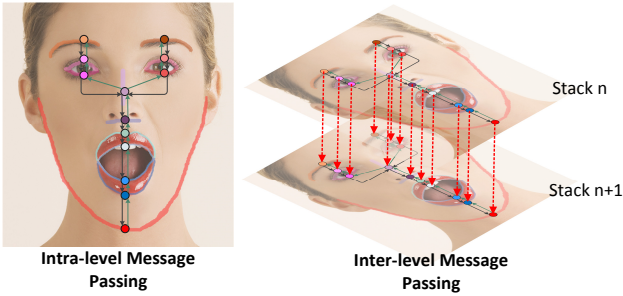

Figure 6: An illustration of message pass scheme. A bi-direction tree structure is used for intra-level message passing. Inter-level message is passed between adjacent stacks from lower to higher.

### 3.3. Boundary effectiveness discriminator

In structured boundary heatmap estimator, mean squared error (MSE) is used as the loss function. However, minimizing MSE sometimes makes the prediction look blurry and implausible. This regression-to-the-mean problem is a well-known fact in the literature of super-resolution [40]. It damages the learning of regression network when bad boundary heatmaps are generated.

However, in our framework, the hard-to-define term "quality" of heatmaps has a very clear evaluation metric. If helping to produce accurate landmark coordinates, the boundary heatmap has a good quality. According to this, we propose a landmark based boundary effectiveness discriminator to decide the effectiveness of the generated boundary heatmaps. For a generated boundary heatmap $\hat{M}$ (all index $i$ such as $\hat{M}_i$ is omitted for the simplicity of notation), denote its corresponding generated landmark coordinates set as $\hat{S}$, the ground-truth distance matric map as $Dist$. The ground truth $d_{\text{fake}}$ of discriminator $D$ that determines whether the generated boundary heatmap is fake can be defined as

$$d_{\text{fake}}(\hat{M}, \hat{S}) = \begin{cases} 0, & \Pr_{s \in \hat{S}}(Dist(s) < \theta) < \delta \\ 1, & \text{otherwise} \end{cases} \quad (4)$$

Where $\theta$ is the distance threshold to ground truth boundary and $\delta$ is the probability threshold. This discriminator predicts whether most generated corresponding landmarks would be close to the ground truth boundary.

Following [9, 10], we introduce the idea of adversarial learning by pairing the boundary effectiveness discriminator $D$ and the boundary heatmaps estimator $G$. The loss of $D$ can be expressed as:

$$\mathcal{L}_D = -(\mathbb{E}[\log D(M)] + \mathbb{E}[\log(1 - |D(G(I)) - d_{\text{fake}}|)]) \quad (5)$$

Where $M$ is the ground truth boundary heatmap. The discriminator learns to predict ground truth boundary heatmap as one while predict generated boundary heatmap according to $d_{\text{fake}}$.

With effectiveness discriminator, the adversarial loss can be expressed as:

$$\mathcal{L}_A = \mathbb{E}[\log(1 - D(G(I)))] \quad (6)$$

Thus, the estimator is optimised to fool $D$ by giving more plausible and high-confidence maps that will benefit the learning of regression network.

The following pseudo-code shows the training process of the whole methods.

---

**Algorithm 1** The training pipeline of the our method.

**Require:** Training image $I$, the corresponding ground-truth boundary heatmaps $M$ and landmark coordinates $S$, the generation network $G$, the regression network $R$ and the discrimination network $D$.

1: **while** the accuracy of landmarks predicted by $R$ in validation set stops **do**
2:     Forward $G$ by $\hat{M} = G(I)$ and optimize $G$ by minimizing $\|\hat{M} - M\|_2^2 + \mathcal{L}_A$ where $\mathcal{L}_A$ is defined in Eq.6;
3:     Forward $D$ by $\hat{d}_{real} = D(M)$ and optimize $D$ by minimizing the first term of $\mathcal{L}_D$ defined in Eq.5;
4:     Forward $D$ by $\hat{d}_{fake} = D(\hat{M})$ and optimize $D$ by minimizing the second term of $\mathcal{L}_D$ defined in Eq.5;
5:     Forward $R$ by $\hat{S} = R(I, \hat{M})$ and optimize $R$ by minimizing $\|\hat{S} - S\|_2^2$;
6: **end while**

---

### 3.4. Cross-Dataset Face Alignment

Recently, together with impressive progress of algorithms for face alignment, various benchmarks have also been released, *e.g.*, LFPW [3], AFLW [28] and 300-W [39]. However, because of the gap between annotation schemes, these datasets can hardly be jointly used. Models trained on one specific dataset perform poorly on recent in-the-wild test sets.

However, introduction of an annotation transfer component [46, 71, 67, 53] will bring new problems. From a new perspective, we take facial boundaries as an all-purpose middle-level face geometry representation. Facial boundaries naturally unify different landmark definitions with enough landmarks. And it can also be applied to help training landmarks regressor with any specific landmarks definition. The cross-dataset capacity is an important by-product of our methods. Its effectiveness is evaluated in Sec. 4.1.

## 4. Experiments

**Datasets.** We conduct evaluation on four challenging datasets including 300W [39], COFW [5], AFLW [28] and WFLW which is annotated by ourself.

*300W* [39] dataset: 300W is currently the most widely-used benchmark dataset. We regard all the training samples (3148 images) as the training set and perform testing on *(i)* full set and *(ii)* test set. *(i)* Full set contains 689 images and is split into common subset (554 images) and challenging subsets (135 images). *(ii)* Test set is the private test-set used for the 300W competition which contains 600 images.

*COFW* [5] dataset consists of 1345 images for training and 507 faces for testing which are all occluded to different degrees. Each COFW face originally has 29 manually annotated landmarks. We also use the test set which has been re-annotated by [19] with 68 landmarks annotation scheme to allow easy comparison to previous methods.

*AFLW* [28] dataset: AFLW contains 24386 in-the-wild faces with large head pose up to $120°$ for yaw and $90°$ for pitch and roll. We follow [72] to adopt three settings on our experiments: *(i) AFLW-Full*: 20000 and 4386 images are used for training and testing respectively. *(ii) AFLW-Frontal*: 1314 images are selected from 4386 testing images for evaluation on frontal faces.

*WFLW* dataset: In order to facilitate future research of face alignment, we introduce a new facial dataset base on WIDER Face [61] named Wider Facial Landmarks in-the-wild (WFLW), which contains 10000 faces (7500 for training and 2500 for testing) with 98 fully manual annotated landmarks. Apart from landmark annotation, out new dataset includes rich attribute annotations, i.e., occlusion, pose, make-up, illumination, blur and expression for comprehensive analysis of existing algorithms. Compare to previous dataset, faces in the proposed dataset introduce large variations in expression, pose and occlusion. We can simply evaluate the robustness of pose, occlusion, and expression on proposed dataset instead of switching between multiple evaluation protocols in different datasets. The comparison of WFLW with popular benchmarks is illustrated in the supplementary material.

**Evaluation metric.** We evaluate our algorithm using standard normalised landmarks mean error and Cumulative Errors Distribution (CED) curve. In addition, two further statistics *i.e.* the area-under-the-curve (AUC) and the failure rate for a maximum error of $0.1$ are reported. Because of various profile face on AFLW [28] dataset, we follow [72] to use face size as the normalising factor. For other dataset, we follow MDM [51] and [39] to use outer-eye-corner distance as the "inter-ocular" normalising factor. Specially, to compare with the results that reported to be normalised by "inter-pupil" (eye-centre-distance) distance, we report our results with both two normalising factors on Table 1.

**Implementation details.** All training images are cropped and resized to $256 \times 256$ according to provided bounding boxes. The estimator is stacked four times if not specially indicated in our experiment. For ablation study, the estimator is stacked two times due to the consideration of time and computation cost. All our models are trained with *Caffe* [24] on 4 Titan X GPUs. Note that all testing images are cropped and resized according to provided bounding boxes *without any spatial transformation* for fair comparison with other methods. For the limited space of paper, we report all of the training details and experiment settings in our supplementary material.

| Method | Common Subset | Challenging Subset | Fullset |
|---|---|---|---|
| Inter-pupil Normalisation | | | |
| RCPR [6] | 6.18 | 17.26 | 8.35 |
| CFAN [69] | 5.50 | 16.78 | 7.69 |
| ESR [7] | 5.28 | 17.00 | 7.58 |
| SDM [57] | 5.57 | 15.40 | 7.50 |
| LBF [38] | 4.95 | 11.98 | 6.32 |
| CFSS [72] | 4.73 | 9.98 | 5.76 |
| 3DDFA [74] | 6.15 | 10.59 | 7.01 |
| TCDCN [70] | 4.80 | 8.60 | 5.54 |
| MDM [51] | 4.83 | 10.14 | 5.88 |
| RAR [55] | 4.12 | 8.35 | 4.94 |
| DVLN [53] | 3.94 | 7.62 | 4.66 |
| TSR [31] | 4.36 | 7.56 | 4.99 |
| **LAB** | **3.42** | **6.98** | **4.12** |
| **LAB+Oracle** | 2.57 | 4.72 | 2.99 |
| Inter-ocular Normalisation | | | |
| PCD-CNN [2] | 3.67 | 7.62 | 4.44 |
| SAN [59] | 3.34 | 6.60 | 3.98 |
| **LAB** | **2.98** | **5.19** | **3.49** |
| **LAB+Oracle** | 1.85 | 3.28 | 2.13 |

Table 1: Mean error (%) on 300-W Common Subset, Challenging Subset and Fullset (68 landmarks).

| Method | AUC | Failure Rate (%) |
|---|---|---|
| Deng *et al*. [14] | 0.4752 | 5.5 |
| Fan *et al*. [16] | 0.4802 | 14.83 |
| DenseReg + MDM [1] | 0.5219 | 3.67 |
| JMFA [15] | 0.5485 | 1.00 |
| **LAB** | **0.5885** | **0.83** |
| **LAB+Oracle** | 0.7626 | 0.00 |

Table 2: Mean error (%) on 300-W testset (68 landmarks). Accuracy is reported as the AUC and the Failure Rate.

### 4.1. Comparison with existing approaches

#### 4.1.1 Evaluation on 300W

We compare our approach against the state-of-the-art methods on 300W Fullset. The results are shown in Table 1. Our method significantly outperforms previous methods by a large margin. Note that, our method achieves 6.98% mean

| Metric | Method | Testset | Pose Subset | Expression Subset | Illumination Subset | Make-Up Subset | Occlusion Subset | Blur Subset |
|---|---|---|---|---|---|---|---|---|
| Mean Error (%) | ESR [7] | 11.13 | 25.88 | 11.47 | 10.49 | 11.05 | 13.75 | 12.20 |
| | SDM [57] | 10.29 | 24.10 | 11.45 | 9.32 | 9.38 | 13.03 | 11.28 |
| | CFSS [72] | 9.07 | 21.36 | 10.09 | 8.30 | 8.74 | 11.76 | 9.96 |
| | DVLN [53] | 6.08 | 11.54 | 6.78 | 5.73 | 5.98 | 7.33 | 6.88 |
| | **LAB** | **5.27** | **10.24** | **5.51** | **5.23** | **5.15** | **6.79** | **6.32** |
| Failure Rate (%) | ESR [7] | 35.24 | 90.18 | 42.04 | 30.80 | 38.84 | 47.28 | 41.40 |
| | SDM [57] | 29.40 | 84.36 | 33.44 | 26.22 | 27.67 | 41.85 | 35.32 |
| | CFSS [72] | 20.56 | 66.26 | 23.25 | 17.34 | 21.84 | 32.88 | 23.67 |
| | DVLN [53] | 10.84 | 46.93 | 11.15 | 7.31 | 11.65 | 16.30 | 13.71 |
| | **LAB** | **7.56** | **28.83** | **6.37** | **6.73** | **7.77** | **13.72** | **10.74** |
| AUC | ESR [7] | 0.2774 | 0.0177 | 0.1981 | 0.2953 | 0.2485 | 0.1946 | 0.2204 |
| | SDM [57] | 0.3002 | 0.0226 | 0.2293 | 0.3237 | 0.3125 | 0.2060 | 0.2398 |
| | CFSS [72] | 0.3659 | 0.0632 | 0.3157 | 0.3854 | 0.3691 | 0.2688 | 0.3037 |
| | DVLN [53] | 0.4551 | 0.1474 | 0.3889 | 0.4743 | 0.4494 | 0.3794 | 0.3973 |
| | **LAB** | **0.5323** | **0.2345** | **0.4951** | **0.5433** | **0.5394** | **0.4490** | **0.4630** |

Table 3: Evaluation of LAB and several state-of-the-arts on Testset and 6 typical subsets of WFLW (98 landmarks).

| Method | Mean Error (%) | Failure Rate (%) |
|---|---|---|
| Human | 5.6 | - |
| RCPR [6] | 8.50 | 20.00 |
| HPM [19] | 7.50 | 13.00 |
| CCR [17] | 7.03 | 10.9 |
| DRDA [68] | 6.46 | 6.00 |
| RAR [55] | 6.03 | 4.14 |
| SFPD [54] | 6.40 | - |
| DAC-CSR [18] | 6.03 | 4.73 |
| **LAB w/o boundary** | 5.58 | 2.76 |
| **LAB** | **3.92** | **0.39** |

| Method | AFLW-Full (%) | AFLW-Frontal (%) |
|---|---|---|
| CDM [64] | 5.43 | 3.77 |
| RCPR [6] | 3.73 | 2.87 |
| ERT [26] | 4.35 | 4.35 |
| LBF [38] | 4.25 | 2.74 |
| CFSS [72] | 3.92 | 2.68 |
| CCL [73] | 2.72 | 2.17 |
| TSR [31] | 2.17 | - |
| DAC-OSR [18] | 2.27 | 1.81 |
| **LAB w/o boundary** | 1.85 | 1.62 |
| **LAB** | **1.25** | **1.14** |

(a) Mean error (%) on COFW-29 testset.  (b) Mean error (%) on AFLW testset.

Table 4: Cross-dataset evaluation on COFW and AFLW.

error on the Challenging subset which reflects the effectiveness of handling large head rotation and exaggerated expressions. Apart from 300W Fullset, we also show our results on 300W Testset in Table 2. Our method performs best among all of the state-of-the-art methods.

To verify the effectiveness and potential of boundary maps, we use ground truth boundary in the proposed method and report results named "LAB+oracle" which significantly outperform all the methods. The results demonstrate the effectiveness of boundary information and show great potential performance gain if the boundary information can be well captured.

### 4.1.2 Evaluation on WFLW

For comprehensively evaluating the robustness of our method, we report mean error, failure rate and AUC on the Testset and six typical subsets of WFLW on Table. 3. These six subsets were split from Testset by the provided attribute annotations. Though reasonable performance is obtained, there is illustrated to be still a lot of room for improvement for the extreme diversity of samples on WFLW, *e.g.*, large pose, exaggerated expressions and heavy occlusion.

### 4.1.3 Cross-dataset evaluation on COFW and AFLW

COFW-68 is produced by re-annotating COFW dataset with 68 landmarks annotation scheme to perform cross-dataset experiments by [19]. Fig. 7 shows the CED curves of our method against state-of-the-art methods on the COFW-68 [19] dataset. Our model outperforms previous results with a large margin. We achieve 4.62% mean error with 2.17% failure rate. The failure rate is significantly reduced by 3.75%, which indicates the robustness of our method to handle occlusions.

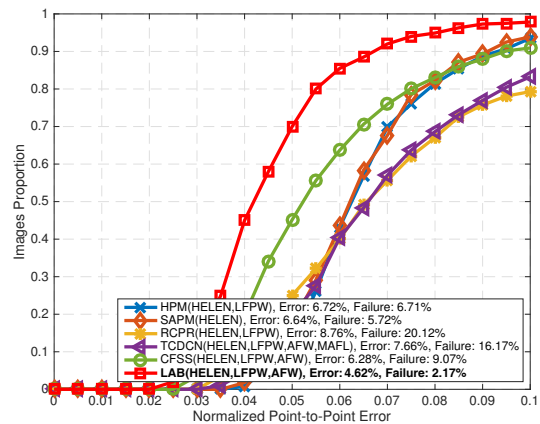

Figure 7: CED for COFW-68 testset (68 landmarks). Train set (in parentheses), mean error and failure rate are also reported.

In order to verify the capacity of handling cross-dataset face alignment of our method, we use boundary heatmaps estimator trained on 300W Fullset which has no overlap with COFW and AFLW dataset and compare the performance with and without using boundary information fusion ("LAB w/o boundary"). The results are reported in Table 4. The performance of previous methods without using 300-W datasets is also attached as a reference. There is a clear boost between our method without and with using boundary information. Thanks to the generalization of facial boundaries, the estimator learned on 300W can be conveniently used to supply boundary information for coordinate regression on COFW-29 [5] and AFLW [28] dataset, even though these datasets have different annotation protocols.

Moreover, our method uses boundary information achieves 29%, 32% and 29% relative performance improve-ment over the baseline method ("LAB without boundary") on COFW-29, AFLW-Full and AFLW-Frontal respectively. Since COFW covers different level of occlusion and AFLW has significant view changes and challenging shape variations, the results emphasise the robustness brought by boundary information to occlusion, pose and shape variations. More qualitative results are demonstrated in our supplementary material.

### 4.2. Ablation study

Our framework consists of several pivotal components, *i.e.*, boundary information fusion, message passing and adversarial learning. In this section, we validate their effectiveness within our framework on the 300W Challenging Set and WFLW Dataset. Based on the baseline res-18 network (BL), we analyse each proposed component, *i.e.*, with the baseline hourglass boundary estimator ("HBL"), message passing ("MP"), and adversarial learning ("AL"), by comparing their mean error and failure rate. The overall results are shown in Fig. 8.

**Boundary information** is chosen as geometric structure representation in our work. We verify the potential of other structure information as well, *i.e.*, facial parts gaussian ("FPG") and face parsing results ("FP"). We report the landmarks accuracy with oracle results in Table 5 using different structure information. It can be observed easily that boundary map ("BM") is the most effective one.

**Boundary information fusion** is one of the key steps in our algorithm. We can fuse boundary information at different levels for the regression network. As indicated in Table 6, our final model that fuses boundary information in all four levels improves mean error from 7.12% to 6.13%. To evaluate the relationship between the quantity of boundary information fusion and the final prediction accuracy, we vary the number of fusion levels from 1 to 4 and report the mean error results in Table 6. It can be observed that performance is improved consistently by fusing boundary heatmaps at more levels.

| Method | BL | BL+FPG | BL+FP | BL+BM |
|---|---|---|---|---|
| Mean Error | 7.12 | 5.25 | 4.16 | **3.28** |

Table 5: Mean error (%) on 300W Challenging Set for evaluation the potential of boundary map as the facial structure information.

| Method | BL | BL+L1 | BL+L1&2 | BL+L1&2&3 | BL+L1&2&3&4 |
|---|---|---|---|---|---|
| Mean Error | 7.12 | 6.56 | 6.32 | 6.19 | **6.13** |

Table 6: Mean error (%) on 300W Challenging Subset for various fusion levels.

| Method | BL | BL+HG/B | BL+CL | BL+HG |
|---|---|---|---|---|
| Mean Error | 7.12 | 6.95 | 6.24 | **6.13** |

Table 7: Mean error (%) on 300W Challenging Set for different settings of boundary fusion scheme.

| Method | HBL | HBL+MP | HBL+MP+AL |
|---|---|---|---|
| Error of heatmap | 0.85 | 0.76 | **0.63** |
| Error of landmark | 6.13 | 5.82 | **5.59** |

Table 8: Normalised pixel-to-pixel error (%) of heatmap estimation, mean error (%) and failure rate (%) of landmark prediction on 300W Challenging Set for evaluation the relationship between the quality of estimated boundary and final prediction.

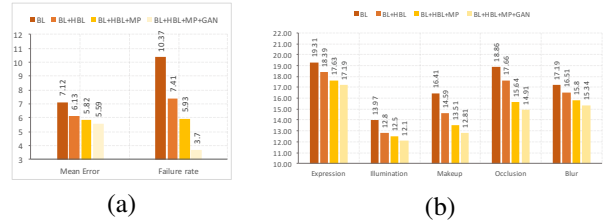

(a)      (b)

Figure 8: (a) Mean error (%) and failure rate (%) on 300W Challenging Subset. (b) Mean error (%) on 5 typical testing subset of WFLW Dataset, *i.e.* Expression, Illumination, Makeup, Occlusion and Blur Subset.

To verify the effectiveness of the fusion scheme shown in Fig. 4, we report the results of mean error on several settings in Table 7, *i.e.*, the baseline res-18 network ("BL"), hourglass module without boundary feature ("HG/B"), hourglass module with boundary feature ("HG") and consecutive convolutional layers with boundary feature ("CL"). The comparison between "BL+HG" and "BL+HG/B" indicates the effectiveness of boundary information fusion rather than network structure changes. The comparison between "BL+HG" and "BL+CL" indicates the effectiveness of the using hourglass structure design.

**Message passing** plays a vital role for heatmap quality improvement when severe occlusions happen. As illustrated in Fig. 8 (b) on Occlusion Subset of WFLW, message passing, which combines information from visible boundaries and occluded ones, reduce the mean error over 11% relatively.

**Adversarial learning** further improves the quality and effectiveness of boundary heatmaps. As illustrated in Fig. 5, heatmaps can be observed to be more focused and salience when adversarial loss is added. To verify the effectiveness of our landmark based boundary effectiveness discriminator, a baseline method using traditionally defined discriminator is tested on 300W Challenging Set. The failure rate is reduced from 5.19% to 3.70%.

**Relationship between boundary estimator and landmarks regressor** is evaluated by analyzing the quality of estimated heatmap and final prediction accuracy. We report the MSE of estimated heatmaps and corresponding landmarks accuracy in Table 8. We observe that with message passing ("HBL+MP") and adversarial learning ("HBL+AL"), the errors of estimated heatmaps are reduced together with landmarks accuracy.

# 5. Conculsion

Unconstrained face alignment is an emerging topic. In this paper, we present a novel use of facial boundary to derive facial landmarks. We believe the reasoning of a unique facial structure is the key to localise facial landmarks, since human face does not include ambiguities. By estimating facial boundary, our method is capable of handling arbitrary head poses as well as large shape, appearance, and occlusion variations. Our experiment shows the great potential of modeling facial boundary. The runtime of our algorithm is $60ms$ on TITAN X GPU.

**Acknowledgement** This work was supported by The National Key Research and Development Program of China (Grand No.2017YFC1703300).